\documentclass[10pt,journal]{IEEEtran}
\usepackage{amsmath,amsfonts}
\usepackage{array}
\usepackage{textcomp}
\usepackage{stfloats}
\usepackage{url}
\usepackage{verbatim}
\usepackage{graphicx}
\usepackage{algorithm}
\usepackage{algpseudocode}
\usepackage{balance}
\usepackage{subcaption}
\usepackage[table]{xcolor}
\usepackage{pifont}
\usepackage{amsmath}
\usepackage{graphicx}
\usepackage{multirow}
\usepackage{tikz-dependency}
\newcommand*\circled[1]{\tikz[baseline=(char.base)]{        
 \node[shape=circle,fill,inner sep=0.5pt] (char) {\textcolor{white}{#1}};}}

\def\BibTeX{{\rm B\kern-.05em{\sc i\kern-.025em b}\kern-.08em
    T\kern-.1667em\lower.7ex\hbox{E}\kern-.125emX}}

\begin{document}

\title{A Generalized Transformer-based Radio Link Failure Prediction Framework in 5G RANs}

\author{
	\IEEEauthorblockN{
		Kazi Hasan, Thomas Trappenberg, Israat Haque}\\
	\IEEEauthorblockA{
		Dalhousie University, Canada}
  }

\maketitle

\begin{abstract}

Radio link failure (RLF) prediction system in Radio Access Networks (RANs) is critical for ensuring seamless communication and meeting the stringent requirements of high data rates, low latency, and improved reliability in 5G networks. However, weather conditions such as precipitation, humidity, temperature, and wind impact these communication links. Usually, historical radio link Key Performance Indicators (KPIs) and their surrounding weather station observations are utilized for building learning-based RLF prediction models. However, such models must be capable of learning the spatial weather context in a dynamic RAN and effectively encoding time series KPIs with the weather observation data. Existing works fail to incorporate both of these essential design aspects of the prediction models. 
This paper fills the gap by proposing \textit{GenTrap}, a novel RLF prediction framework that introduces a \textit{graph neural network (GNN)}-based learnable weather effect aggregation module and employs state-of-the-art time series \textit{transformer} as the temporal feature extractor for radio link failure prediction. The proposed aggregation method of GenTrap can be integrated into any existing prediction model to achieve better performance and generalizability. We evaluate GenTrap on two real-world datasets (rural and urban) with 2.6 million KPI data points and show that GenTrap offers a significantly higher F1-score (0.93 for rural and 0.79 for urban) compared to its counterparts while possessing generalization capability.

\end{abstract}



\section{Introduction}

The emergence of modern networking applications such as Industry 4.0, intelligent transportation systems, health informatics, and augmented/virtual reality (AR/VR) necessitates high network bandwidth, robust reliability, and fast communication speeds \cite{ding2018opportunities}. Fifth-generation (5G) cellular networks aspire to accommodate these applications by satisfying various service level objectives (SLOs) through the use of millimetre-wave (mmWave) spectrums (24GHz to 100GHz). 
For instance, teleoperated driving systems can schedule human takeover if SLOs are not satisfied \cite{boban2021predictive}. However, a 5G radio access network (RAN) requires deploying a denser array of base stations to communicate over mmWave radio as it traverses short distances. Moreover, these links suffer from distortion and attenuation due to weather phenomena like precipitation, humidity, temperature, and wind \cite{saadoon2021overview,abuhdima2021impact}. Thus, mobile operators must have predictive maintenance of RANs connections to support the above real-time applications. 

A group of works investigated this correlation of mmWave radio and weather conditions using operator-provided real data \cite{agarwal2022prediction, AKTAS2022107742, Islam2022}. Our initial investigation on the same dataset \cite{AKTAS2022107742} reveals the same findings and motivates us to revisit the current reliable 5G RAN systems. These works developed learning-based failure prediction schemes due to the availability of radio key performance indicators (KPIs) from radio stations and respective weather attributes. The automated systems also reduce human intervention, CAPEX, and OPEX. Semih \textit{et al.} propose a branched LSTM architecture to process both temporal and spatial features and offer better performance compared to models considering only temporal data processing \cite{AKTAS2022107742}. Islam \textit{et al.} introduce a comprehensive data preprocessing pipeline and use an LSTM-autoencoder model to predict link failures based on the reconstruction loss \cite{Islam2022}. Finally, Agarwal \textit{et al.} deploy decision trees and random forest classifiers to predict upcoming five-day radio link failures \cite{agarwal2022prediction}. 

These works demonstrate the influence of weather attributes on radio communication and incorporate them into their developed models. However, the solutions suffer from few critical limitations. Specifically, these models cannot capture long-term dependencies due to vanishing gradients issue of LSTM \cite{zhao2020rnn}. Also, they do not weigh the importance of different elements in a sequence during predictions \cite{vaswani2017attention}. Another critical design aspect is correctly associating each link to the surrounding weather stations that affect the link. However, the existing approaches deploy heuristics to associate radio sites with weather stations. For instance, \cite{agarwal2022prediction}, \cite{AKTAS2022107742}, and \cite{Islam2022} use the closest weather station, the aggregated $k$ nearest stations, and the maximum of minimum distances between radio and weather sites, respectively. Also, these heuristic-based solutions cannot generalize well on a dynamic topology, which may occur if 
providers selectively turn on/off radio stations for efficient resource utilization \cite{oh2013dynamic}. Overall, existing LSTM-based solutions suffer from \textit{learnability} and {generalizability} critical for predictive RAN maintenance.  

This work fills the above gaps and proposes \textit{GenTrap}, a novel radio link failure prediction framework capable of efficiently learning both the \textit{spatial} (radio and weather stations association) and \textit{temporal} (time-series features) context of RAN and surrounding weather stations. Specifically, GenTrap leverages a Graph Neural Network (GNN) to dynamically learn the effect of weather attributes from surrounding stations on radio links and realize a generalized model for unseen radio links. This GNN module can be incorporated into existing architectures for better performance and generalizability. In addition, the time series Transformer can encode complex temporal dependencies by learning which time point in the past to focus on for predicting future link failures \cite{zerveas2021transformer}. Thus, the novelty of GenTrap includes developing a learnable architecture for capturing weather effects and applying transformers in encoding time series radio and weather data for radio link failure prediction.

The GNN weather effect learnable module dynamically assigns $k$ nearest weather stations to each radio link, i.e., allows each link to learn from different values of $k$. Then, the time series transformer module encodes each pair of historical radio link KPI and the assigned weather station observations.
An average pooling operation on the transformer output captures the temporal dependencies among features for each link. A GNN max aggregation over these vectors generates latent representations for spatio-temporal correlations between links and their associated weather stations. The static features of the link are processed using a feed-forward network. Finally, the feed-forward output is concatenated with the spatio-temporal latent vector and passed through a final feed-forward network for link failure prediction the next day. We evaluate GenTrap over two sets (urban and rural) of open-source real-world data provided by the top Turkish telecommunications company, Turkcell \cite{TurkcellDataset,AKTAS2022107742}. The datasets consist of KPIs from both regular and failed links and surrounding weather stations (details in Section~\ref{sec:dataset}). The contributions of this paper are the following:

\begin{itemize}
    \item We propose the GenTrap framework that introduces a generalized weather effect learnable GNN aggregation module and incorporates the state-of-the-art time-series transformer to capture spatio-temporal context efficiently.

    \item To the best of our knowledge, we are the first to propose a novel aggregation module that can also improve existing deep learning-based failure prediction models. 
    
    \item We rigorously experiment using two real-world data sets to show the superiority of GenTrap in performance and generalizability compared to its counterpart. Specifically, it offers an F1 score of 0.92 and improves the F1 score of LSTM+ from 0.63 to 0.70, incorporating the proposed aggregation module.      

    \item We share the GenTrap prototype code \cite{gentrap} for reproducing, adapting, and extending the proposed framework.
\end{itemize}

\section{Background and Motivation}

This section offers the necessary background to understand the propose work and associates that knowledge to motivate the need for developing GenTrap. 

\begin{figure}[h]
    \begin{center}
        \centering
        {\includegraphics[width=3.15 in]{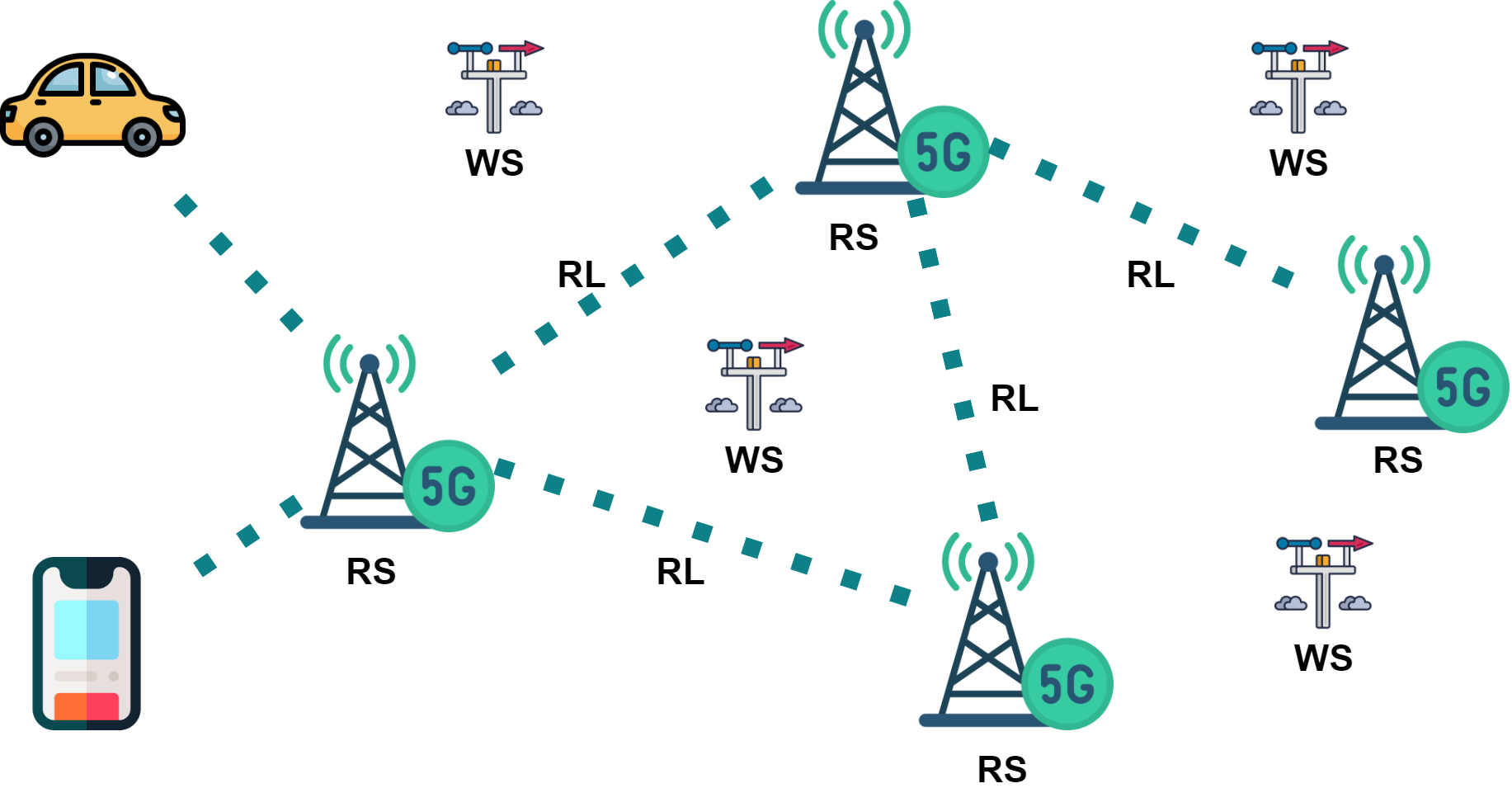}}
        \caption{ An example of 5G RAN deployment. }
        \vspace{-0.2in}
        \label{fig:deployment}
    \end{center}
\end{figure}

\textbf{5G RAN.} 5G has gained popularity due to its support for emerging applications that require high bandwidth, high reliability, and low latency \cite{Guevara2020Roleof5G}. It uses mmWaves to achieve these breakthroughs at the cost of coverage size and higher penetration loss \cite{Zekri2020AnalysisofOutdor}. 5G deployment takes advantage of small cells that collect user traffic and communicate it to radio sites/stations (RS) to mitigate these drawbacks. The radio sites use 5G radio links (RL) to communicate with each other and the core network, which connects users to the Internet \cite{Habibi2019ComprehensiveSurveyRan}. These links are usually surrounded by numerous weather stations (WS) that can provide weather context at the radio links \cite{AKTAS2022107742}. An overview of this type of deployment is depicted in Fig.~\ref{fig:deployment}. We use radio site KPIs along with the weather station data to predict radio link failures for the upcoming day. Thus, providers can take the necessary precautions for critical services. 

\textbf{Imbalance dataset.} The percentage of radio link failures, however, is less compared to the normally operating ones, which creates imbalanced data to be processed. For instance, the Turkcell dataset \cite{AKTAS2022107742} that we use has 0.3\% and 0.06\% failures in rural and urban deployments, respectively \cite{AKTAS2022107742}. Thus, deep learning models trained on such a highly imbalanced dataset leads to good performance only on the majority class \cite{Johnson2019}. Typical approaches for handling such datasets are to use random undersampling of the majority class \cite{AKTAS2022107742} and SMOTE oversampling of the minority class \cite{Islam2022}. Undersampling balances class distributions by randomly removing majority class instances, which may remove informative data points \cite{Mohammed2020}. On the other hand, SMOTE oversampling takes each minority class sample and generates synthetic examples along the line segments by joining $k$ nearest neighbours \cite{Chawla2002SMOTE}. However, some of the limitations of SMOTE include generating noisy samples and introducing bias due to sub-optimal neighbours selection \cite{Jiang2021OversamplingBasedOnContribution}. 


We use the weighted cross entropy loss function to overcome the limitations of undersampling and oversampling as data points are neither sampled nor generated \cite{Rezaei2020AddressingImbalanceinMulti,Aurelio2019LearningFromImbalanced,Ibomoiye2021CostSensitiveLearning}. It deals with the imbalanced data by incorporating prior probabilities into a cost-sensitive cross-entropy error function \cite{Aurelio2019LearningFromImbalanced}. The regular cross-entropy function is symmetrical. Also, the error reduction for both classes occurs at the same logarithmic rate. Thus, in the case of imbalanced data, the majority class will have a larger influence on the total loss as the overall error is minimized regardless of class. We use the following weighted cross entropy loss function to mitigate the effects of class imbalance in the dataset. 

\begin{equation}
    J(\theta) = \frac{1}{m} \sum_{i=1}^{m} -y^i\log(\hat{y}^i)(1-\lambda) - (1-y^i)\log(1-\hat{y}^i)\lambda \label{eq:weighted}
\end{equation}
Here, $J(\theta)$ and $m$ are the total loss and number of samples, respectively. $y$ is the ground truth ($y=1$ for failure) and $\hat{y}$ is the model prediction. 

\textbf{Graph Neural Network (GNN) aggregation.} 
We can deploy a learning-based failure prediction scheme on the balanced data by incorporating weather impact on radio links. For instance, existing works incorporate surrounding weather station information - by using derived features, optimal distance, and closest station data to capture their spatial context \cite{AKTAS2022107742,Islam2022,agarwal2022prediction}. However, the optimal number of closest weather stations can vary across links even in the same deployment; thus, using a fixed number of surrounding weather stations is ineffective. A learning-based approach that can dynamically incorporate and learn the context of surrounding stations can better capture the impact.  
Also, giving surrounding stations the same weight may introduce bias in the prediction as their impact must be proportional to their distance to the given radio station. Thus, we must deploy deep learning techniques that can learn the weighted aggregation of surrounding stations. Finally, 5G deployments are different from urban (denser) to rural (sparser) and also within a single deployment (e.g., patches of dense regions); as a result, algorithms which do not extract all useful information fail to generalize over new radio links. Thus, a generalized algorithm should have robust representation learning capabilities to encode useful context for each radio link. 


To tackle the above issues, we introduce a dynamic weather effect aggregation scheme that uses graph neural networks (GNNs) \cite{HmiltonNIPS2017,xu2021deep,yang2022stam,corso2020principal} to perform feature aggregation and create useful node embeddings from a variable number of neighbors. \textit{Node embedding} compresses high dimensional data of a node's neighborhood into a low dimensional vector embedding to feed into neural networks for classification, clustering, or prediction. A popular GNN model, GraphSAGE is a general framework that aggregates the features from a local neighborhood, and often generalizes better \cite{HmiltonNIPS2017}. Formally, the graph aggregation step can be described by Equation~\ref{eq:gnn}.
\begin{equation}\label{eq:gnn}
\begin{split}
    \{e_{l'},l'\in N(l)\} &= \sigma(W.\{z_{l'},l'\in N(l)\}) \\
    e_l &= max(\{e_{l'},l'\in N(l)\})
\end{split}
\end{equation}
Here, $N(l)$ is the set of neighboring nodes for $l$ and $z_{l'}$ is the set of feature representations of these nodes. A transformation by weights $W$ (can be any neural network) and non-linear function $\sigma$ generates the set of learned feature vectors $\{e_{l'},l'\in N(l)\}$ of the neighboring nodes. Finally, a $max$ operation over these vectors produces the aggregated embedding $e_l$ for the node $l$.


\textbf{Time-series transformer.} 
There is also a temporal aspect of the link failure prediction as both radio link KPI and weather observations are time series data. In recent years, transformer-based time series representation learning models have become popular \cite{Johnson2019}, which are based on a multi-head attention mechanism \cite{Johnson2019,vaswani2017attention} suitable for time-series data \cite{wen2022transformers}. The self-attention module learns to simultaneously represent each element in the input sequence by considering its complete context (dependencies with other elements in the sequence) \cite{vaswani2017attention}. On the other hand, multiple attention heads can consider different representation contexts \cite{zerveas2021transformer}, i.e., multiple types of relevance between input elements in the time-series sequence, which may correspond to multiple kinds of periodicities in the multivariate data. 


\begin{figure}[h]
    \centering
    {\includegraphics[width=2.9 in]{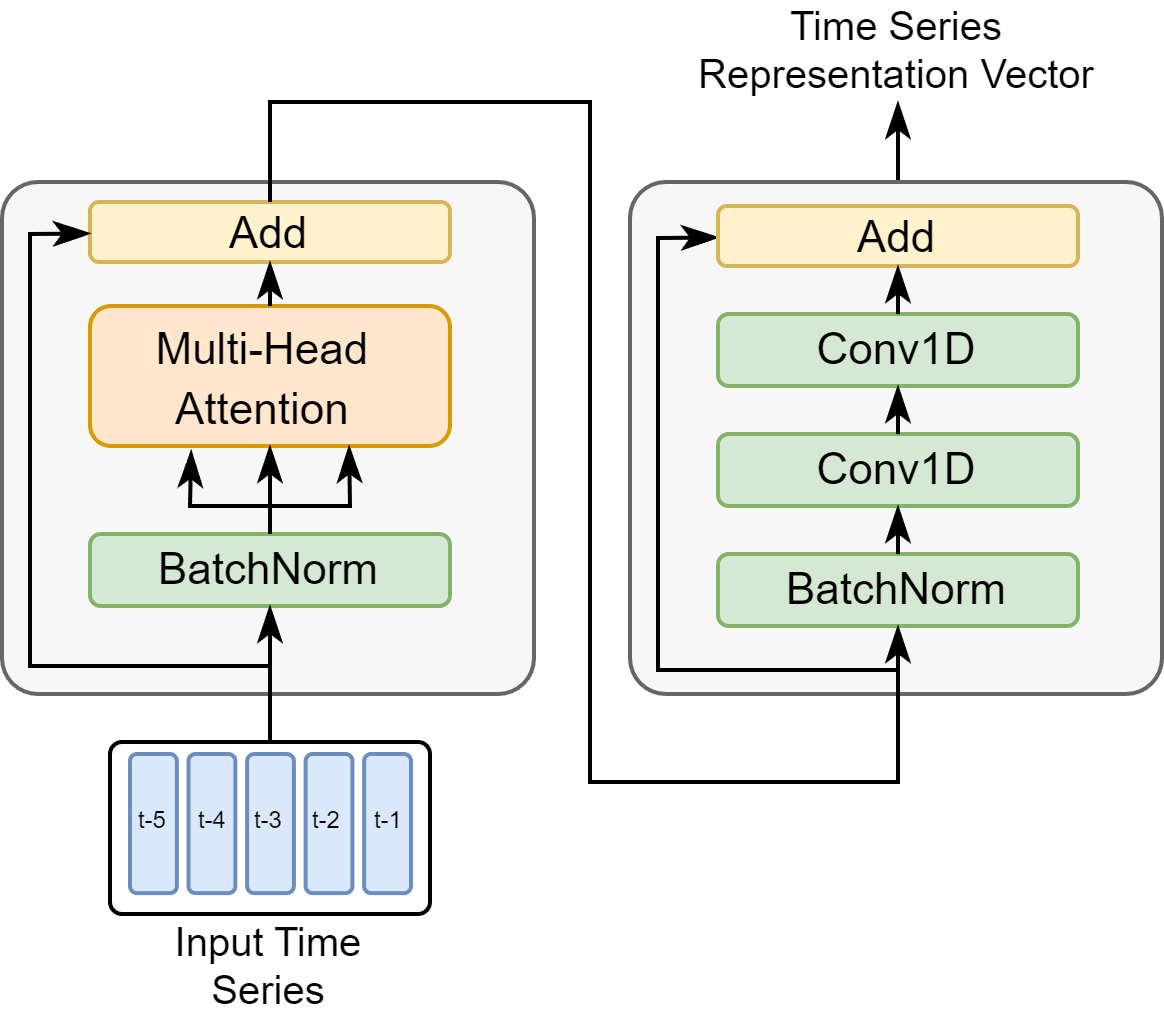}}
    \caption{Transformer module for time-series representation learning. }
    \label{fig:time_trans}
\end{figure}

Fig.~\ref{fig:time_trans} presents the architecture of a time-series transformer module with two sub-modules. The first operates on the time series input sequence and performs batch normalization across the feature dimension. Then, a multi-head attention mechanism \cite{vaswani2017attention} jointly attends to information from different representation sub-spaces at different positions. Residual connections around each of the two sub-modules prevent the vanishing gradient issues, which are absent in LSTM architecture. The second sub-module has one batch normalization layer along with two 1D convolution layers with Relu activation in between that extracts local patterns and features of the sequence. The output time series representation vectors are compressed vector representations of the original sequence that can be fed to downstream tasks, such as classification and regression.

\section{Related work} \label{sec:related-work}

This section presents two groups of related works for GenTrap: learning-based failure prediction approaches in 5G and GNN-based aggregation methods that capture spatial correlations.

\textbf{Learning-based failure prediction.} Khunteta et al. \cite{khunteta2017deep} and Boutiba et al. \cite{boutiba2021radio} introduced the LSTM network to capture temporal feature correlations to predict link failures, but they do not consider weather effects. Other works filled the gap and utilized both historical radio link KPIs and weather observation data - similar to the dataset used in our approach. For example, Agarwal et al. \cite{agarwal2022prediction} combined individual link features with the closest weather station measurements and proposed Random Forest as the classifier. Aktas et al. \cite{AKTAS2022107742} utilized a branched architecture with LSTM and feed-forward network to capture temporal and categorical feature dependencies, respectively. Islam et al. \cite{Islam2022} exploited the advantage of the reconstruction capabilities of LSTM-autoencoder by training their model on normal operational data and flagging data points with high reconstruction error as a failure during testing. These approaches rely on LSTM's ability to extract useful information. Still, they cannot weigh the importance of elements in a time series sequence. They fail to capture all possible influences among time series variables \cite{vaswani2017attention}. Recently, transformer models have demonstrated promising results in time series forecasting \cite{wen2022transformers}, as they can capture long-range dependencies, focus on important elements in a sequence and learn from all possible dependencies \cite{zerveas2021transformer}. We take advantage of the time-series transformer model and propose a branched architecture that performs graph aggregation over each link's surrounding weather stations to achieve the best performance.

\textbf{GNN aggregation to capture spatial correlations.} Effectively capturing spatial dependencies of surrounding weather station data for a radio link is an important step in RLF prediction. Researchers have successfully used GNN based aggregation method in different applications, such as Wu et al. \cite{wu2022inductive} captured spatio-temporal relationships of weather radar for precipitation forecasting, Fan et al. \cite{fan2022gnn} aggregated weather data for crop yield prediction, and Gao et al. \cite{gao2022interpretable} encoded weather parameters for solar radiation prediction. On the other hand, previous works on RLF prediction using weather station data used simple heuristics as part of the data pre-processing step to incorporate spatial relations of weather stations. Agarwal et al. \cite{agarwal2022prediction} only combined the closest weather station features with each radio link, Aktas et al. \cite{AKTAS2022107742} calculated derived features from a fixed $k$ nearest weather stations, and Islam et al. \cite{Islam2022} calculated an optimal distance within which all weather stations were associated with a radio link. These works fail to consider the dynamic relationship between radio links and weather stations as they only consider a fixed number of weather stations for all the radio links and give equal weights to all associated weather stations for a link. We address these shortcomings by using GNN aggregation to capture spatial relationships from surrounding weather stations, where a variable number of weather stations is considered for each radio link and a max aggregation is used. Thus, the model also gains regularization ability along with its better performance.

\section{Dataset Description} \label{sec:dataset}

The performance of GenTrap is evaluated over two sets of real-world open-source data from a renowned telecommunication provider, Turkcell \cite{AKTAS2022107742}. The dataset comprises radio link configuration and key performance indicator (KPIs) data, coupled with time-aligned weather station observations of two distinct deployments, urban and rural, where the time range is between January 2019 to December 2020 and January 2019 to December 2019, respectively. Because of privacy concerns, some configuration parameters and performance data (e.g., equipment name, link IDs, etc.) of the radio links are anonymized without loss of information. Also, the actual GPS location of these stations is not provided; instead, there are pairwise relative distances among these sites. A detailed description of the data tables is provided below.

\textbf{rl-sites.} This data contains radio site identifiers and site-specific parameters such as height and clutter class - surrounding environment at the site, e.g., open urban, open land, dense tree area, etc. The same radio site can have multiple radio links as each uses different links to communicate with different sites.

\textbf{rl-kpis.} Presents daily radio link KPIs, where important ones include severally error second, error second, unavailable second, block bit error, etc. and link-specific configuration parameters such as card type, modulation, frequency band, etc. Each link is uniquely identified with a pair of radio site-id and mini link-id. 

\textbf{met-stations.} This data encompasses unique weather station numbers and station-specific parameters such as height and clutter
class information. The surrounding environment can take clutter class values such as dense trees, open land, airports, etc. These features provide the spatial characteristics of the weather stations.

\textbf{met-real.} Provides hourly historical weather observations (e.g., temperature, humidity, precipitation, etc.) that are aggregated daily to align with the radio link KPI data. These features capture the temporal properties of the weather stations.

\textbf{met-forecast.} This data provides the upcoming five-day weather forecast (snow, rain, scattered clouds, etc.), humidity, temperature, wind speed, etc. for each day. The maximum and minimum predictions are also provided for forecast features (e.g., temperature and humidity).  

\textbf{distances.} Contains pairwise relative distances between all radio sites and weather stations, where the distances are considered in units.

The above datasets contain similar features in both urban and rural deployments but with different numbers of radio sites, radio links, and weather stations (Table \ref{tab:data}). 

\begin{figure}[h]
    \centering
    {\includegraphics[width=2.9 in]{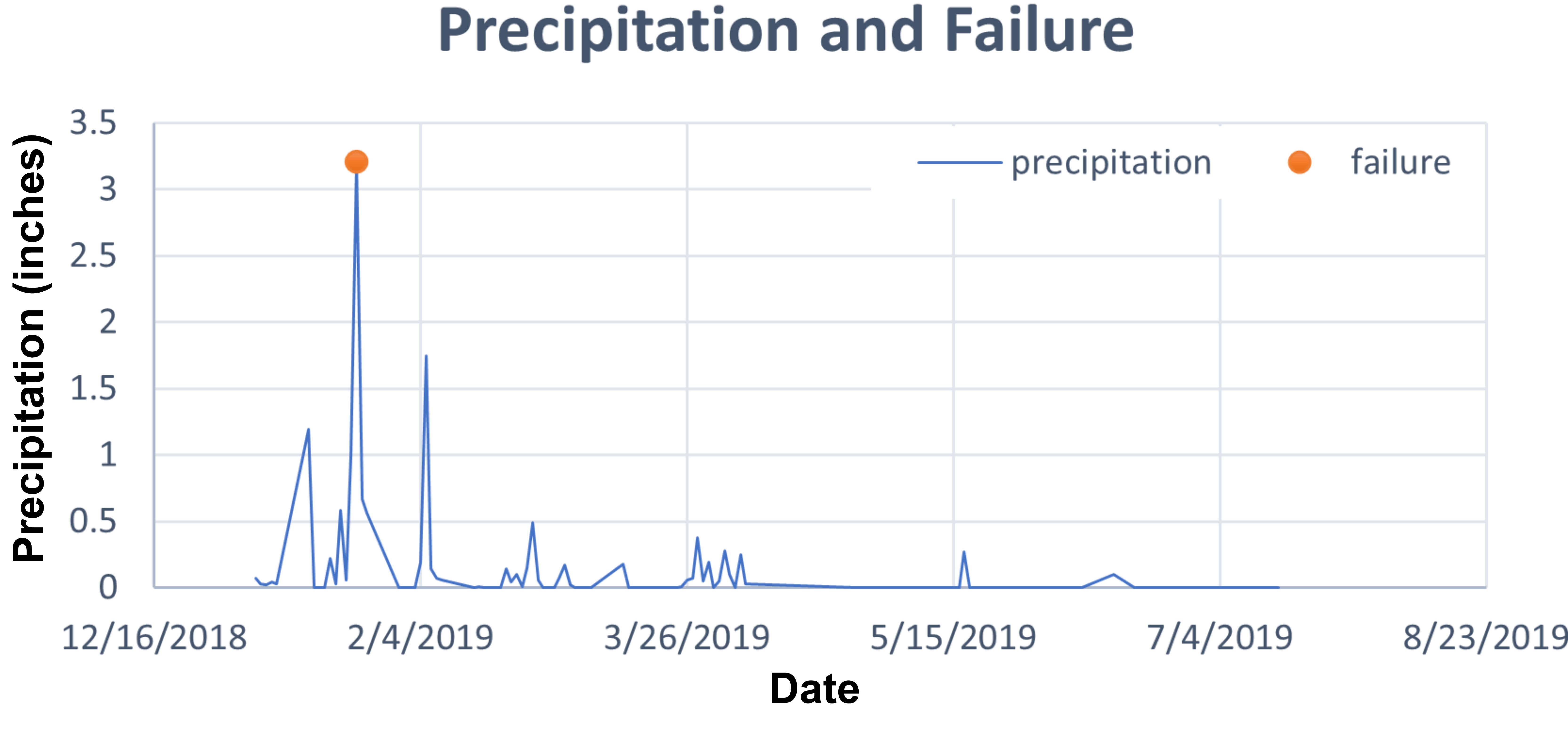}}
    \caption{Impact of precipitation on radio link failure.}
    \label{fig:weather_impact}
\end{figure}

\textbf{Impact of weather on radio links.} We try to understand the relationship between surrounding weather station data and radio link failure through an initial investigation. In Fig \ref{fig:weather_impact}, we plot precipitation recorded from the closest weather station of a radio link and observe that failure occurs during peak rainfall. This demonstrates heavy precipitation has impact on 5G communication channels.

\begin{table}[!t]
\centering
\caption{Summary of the Dataset.}
\label{tab:data}
\begin{tabular}{|p{4.0cm}|p{1.5cm}|p{1.5cm}|} 
 \hline
  & Urban & Rural \\ [0.5ex] 
 \hline\hline
 Number of radio sites & 1674 & 1674 \\ 
 \hline
 Number of weather stations & 20 & 117 \\
 \hline
 Number of time-series radio link KPI features & 7 & 7 \\
 \hline
 Number of time-series weather features & 9 & 9 \\
 \hline
 Total sample size & 1.8 million & 0.4 million \\
 \hline
\end{tabular}
\end{table}

\section{GenTrap Architecture} \label{sec:method}

This section first presents the GenTrap architecture. Then, we illustrate the integration of GNN-based spatial context capturing in existing LSTM+ and LSTM-Autoencoder models. 

\subsection{GenTrap}

\begin{figure*}[!t]
    \begin{center}
        \centering
        {\includegraphics[width=7 in]{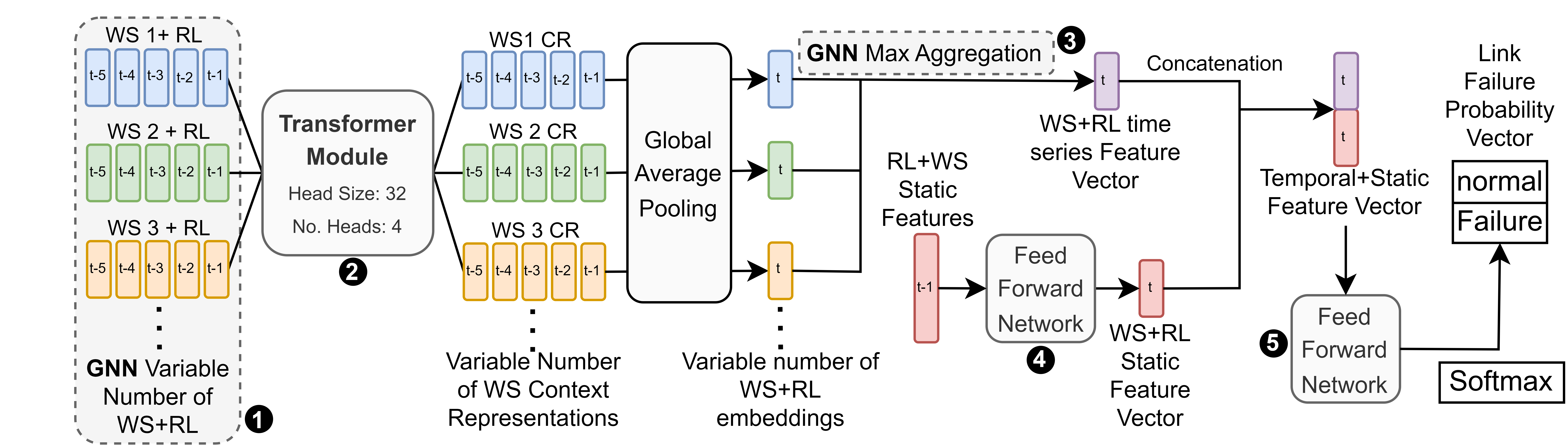}}\\
        \caption{GenTrap architecture.}
        \label{fig:architecture}
    \end{center}
\end{figure*}

Fig.~\ref{fig:architecture} presents the GenTrap architecture, which maps time-series sequences to a probability vector. Specifically, the prediction system takes the radio link KPIs and surrounding weather station observations as inputs and generates the link failure probability vector for the following day. The system deploys \circled{1} GNN aggregation over variable weather stations, \circled{2} a time-series transformer, and \circled{3} a GNN max aggregation function to generate an embedding vector that captures the given radio link and its relevant weather station context. In parallel, \circled{4} a feed-forward network processes one hot encoded static feature and generates a latent representation to capture link and weather station configuration parameters. Then, these two context feature vectors are concatenated and fed to \circled{5} another feed-forward network to produce the final output vector with two elements: the probability of link failure on the following day and the probability of no failure. We divide GenTrap into three main modules: GNN aggregation, generalized transformer branch, and feed-forward branches, which are presented below.

\subsubsection{Learnable GNN Aggregation} 

This is the key component of GenTrap in learning the spatial context among radio and weather stations to realize a generalized model. The GNN aggregation scheme is presented in Algorithm~\ref{alg:cap} that has two components (highlighted in Fig. \ref{fig:architecture}): \circled{1} \textit{GNN Variable Number of $WS+RL$} and \circled{3} \textit{GNN Max Aggregation}. In the first component, for each mini-batch of $m$ links, we pick a value $k$ where $k$ can range from one to the maximum number of closest weather stations {(Line~\ref{line:k_rand}). Thus, links in mini-batches can be associated with different numbers of closest weather stations. Specifically, we consider a variable number of weather stations for each link {(Line~\ref{line:GNNvar})}. We iterate over its $k$ closest weather stations {(Line~\ref{line:k_iter})} and concatenate the KPI feature vector with the corresponding time-aligned weather station observation vectors to generate $k$ $WS+RL$ vectors {(Line~\ref{line:concat})} for the chosen radio link. Then, we use the Transformer module (presented below) to convert them into context-aware representation vectors {(Line~\ref{line:trans})}. In the second component, we take this output and perform the global average pooling to create $k$ temporal embedding vectors that capture the time series dependencies over historical link and weather data. We do a max aggregation {(Line~\ref{line:max})} across these vectors to get our final node embedding vector ($L^mNodeEmbd$) for a radio link. Similarly, we calculate node embedding vectors for all $m$ links.


\begin{algorithm}
\caption{Weather Station Aggregation}\label{alg:cap}
  \textbf{Input:} Historical Radio Link KPIs of $l$ links for $t$ days, where each link $L=\{L^1,L^2,L^3,...,L^l\}$, and $L\in R^{lXtXfeatures}$; Historical weather station observations of $n$ stations for $t$ days, where $W=\{W^1,W^2,W^3,...,W^n\}$, and $W \in R^{nXtXfeatures}$; Transformer weight matrices $T$; Differentiable aggregator function $max$; $M$ mini batches with each of size $m$. \\
 \textbf{Output:} Node embeddings for all links in a mini batch
\begin{algorithmic}[1]
\For{$minibatch \gets 1$ to $M$}
    \State $k \gets Random [1,n]$ \label{line:k_rand}
    \For{$L^m \gets 1$ to $m$}
        \State $EmbdList \gets \emptyset$
        \State {\# (GNN Variable WS input)} \label{line:GNNvar}
        \For{$W^k \gets 1$ to $k$ closest stations} \label{line:k_iter}
            \State $WS+RL \gets concat(L^m,W^k)$ \label{line:concat}
            \State $ConReps \gets T(WS+RL)$ \label{line:trans}
            \State $TempEmbd \gets AvgPooling(ConReps)$
            \State $ EmbdList \gets $ \textbf{append} $TempEmbd$
        \EndFor
        \State {\# (GNN Max Aggregation)} \label{line:GNNmax}
        \State $L^mNodeEmbd \gets max(EmbdList)$ \label{line:max}
    \EndFor
\EndFor
\end{algorithmic}
\end{algorithm}


\subsubsection{Transformer Branch} 

The transformer module goes over the time series vectors as part of the GNN variable weather station consideration process and generates radio link plus weather station embedding vectors that capture the influence of each element on every other element of the time series sequence. 

Specifically, it receives radio link (RL) and weather station (WS) time-series data as input. The RL time series includes $9$ features (e.g., severe error seconds, available time, bbe, etc.), whereas the weather station time series includes $7$ features (e.g., temperature, humidity, precipitation, etc.). These features are available for each day, and we add the time step as an additional feature as a positional encoding scheme for the transformer. We can describe time series vectors for a radio link as $L=\{L_1,L_2,L_3,...,L_t\}$, where $L\in R^{tX9}$ and $t$ ranges from $1$ to the total number of days. While weather station vectors can be represented as $W=\{W^1,W^2,W^3,...,W^n\}$ in ascending order of distance from $L$, where $n$ is the total number of weather stations in a deployment. Each weather station has a time-series data, $W^1=\{W^1_1,W^1_2,W^1_3,...,W^1_t\}$; where $W \in R^{nXtX7}$. 

In Fig.~\ref{fig:architecture}, $WS1+RL$ represents the previous five-day feature vectors of one radio link and its first closest weather station time-series data, whereas $WS2+RL$ represents the same radio link and second closest weather station data. These (WS+RL) vectors are the concatenation of $9$ link and $7$ weather features along with $1$ time-step number column, giving us the input tensor to transformer module, which is of shape {$= (batch size,3X5X17)$}. The transformer module has a multi head attention with $4$ heads, each of size $32$, and the two 1D convolution filters are of size $32$ and $17$, respectively (Fig. \ref{fig:time_trans}). The output shape of the transformer is $(batch size,3X5X17)$, the same as the input shape. We perform global average pooling across the time dimension to capture temporal dependencies for each $WS+RL$ pair, giving us output $(batchsize,5,17)$ - denoted by different colors for different pairs - for each pair. Lastly, we choose the max function as our aggregator to perform an element-wise max operation across the embedding vectors to get one feature vector $(batchsize,17)$. Thus, the generalized transformer module generates one feature vector to capture the variable number of closest weather station effects on each radio link.

\subsubsection{Feed forward Branches} 

We use the generalized transformer module to handle time series radio links and weather station data. On the other hand, we deploy a feed-forward network to encode static radio links and weather station features (Fig. \ref{fig:architecture}). Radio link features (e.g., modulation type, frequency band) and weather station features (e.g., clutter class, weather day) are categorical features. We perform one hot encoding of these categorical features and pass them to a feed-forward network with two layers: $32$ and $17$ neurons. 

The output vector from the static branch is concatenated with the output vector from the generalized transformer branch to get a representation vector $(batchsize,34)$ that captures both temporal and static dependencies. We feed the concatenated vector to another feed-forward network with two layers: $16$ and $2$ neurons. A Softmax layer gets the final probability vector for link failure. We train our model with weighted categorical cross-entropy loss and Adam optimizer. We take the maximum of the $2$ probability scores during inference to make a binary prediction for each input.

\subsection{LSTM+}

Fig.~\ref{fig:lstmp} presents the LSTM+ architecture, and Fig.~\ref{fig:lstmp_g} shows the LSTM+ with augmented GNN-based dynamic weather station aggregation. The LSTM+ architecture uses two separate branches to process time series and static features. The temporal dependencies of radio links and derived weather station features are captured using $4$ LSTM layers. In contrast, the configuration parameters are one-hot encoded and processed by a feed-forward network similar to the ones used in GenTrap. These two output vectors are concatenated and fed to another feed-forward network to get the final probability score vector.

\begin{figure}[h]
    \centering
    \includegraphics[width=0.46\textwidth]{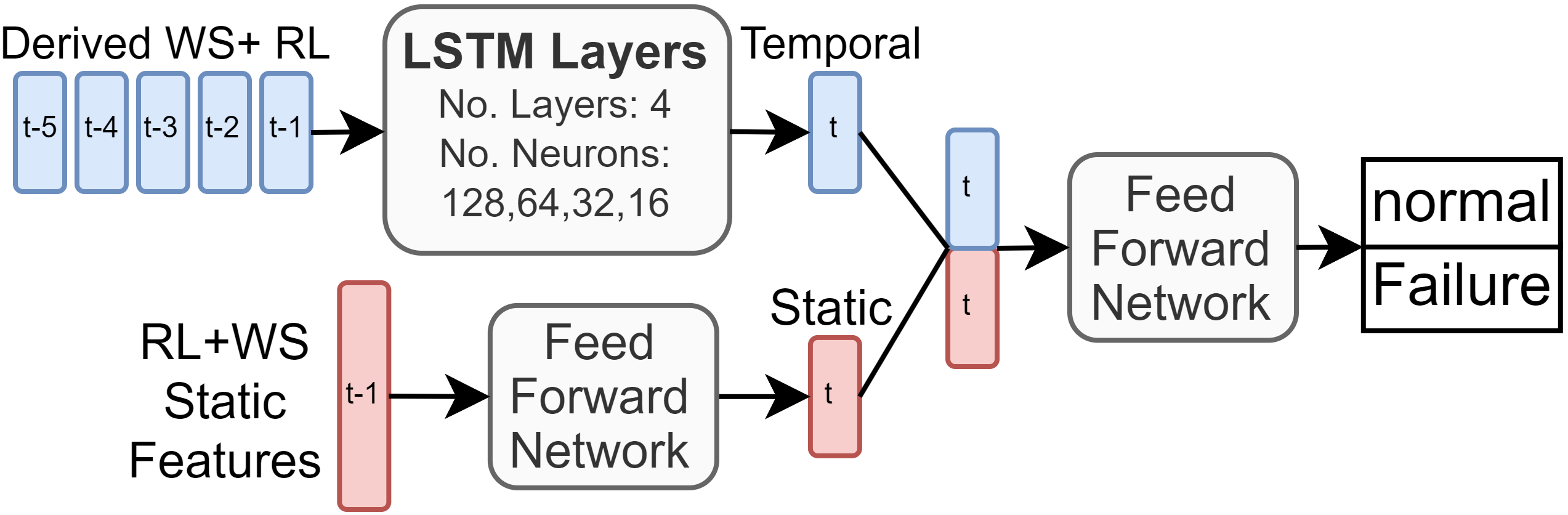}
    \caption{LSTM+ architecture. }
    \label{fig:lstmp}
\end{figure}

We also augment LSTM+ architecture with our generalized graph aggregation method to measure the performance improvement gained by our framework. The GNN augmented LSTM+ architecture incorporates the two components of learnable GNN aggregation (Fig. \ref{fig:lstmp_g}), which generates a variable number of feature representations to capture temporal dependencies of radio link and weather station pairs. A max aggregation function aggregates the representation vectors to capture the spatial correlations among the surrounding weather stations. The static branch and output branches uses feed forward networks that are similar to our proposed approach. 

\begin{figure}[h]
    \centering
    \includegraphics[width=0.46\textwidth]{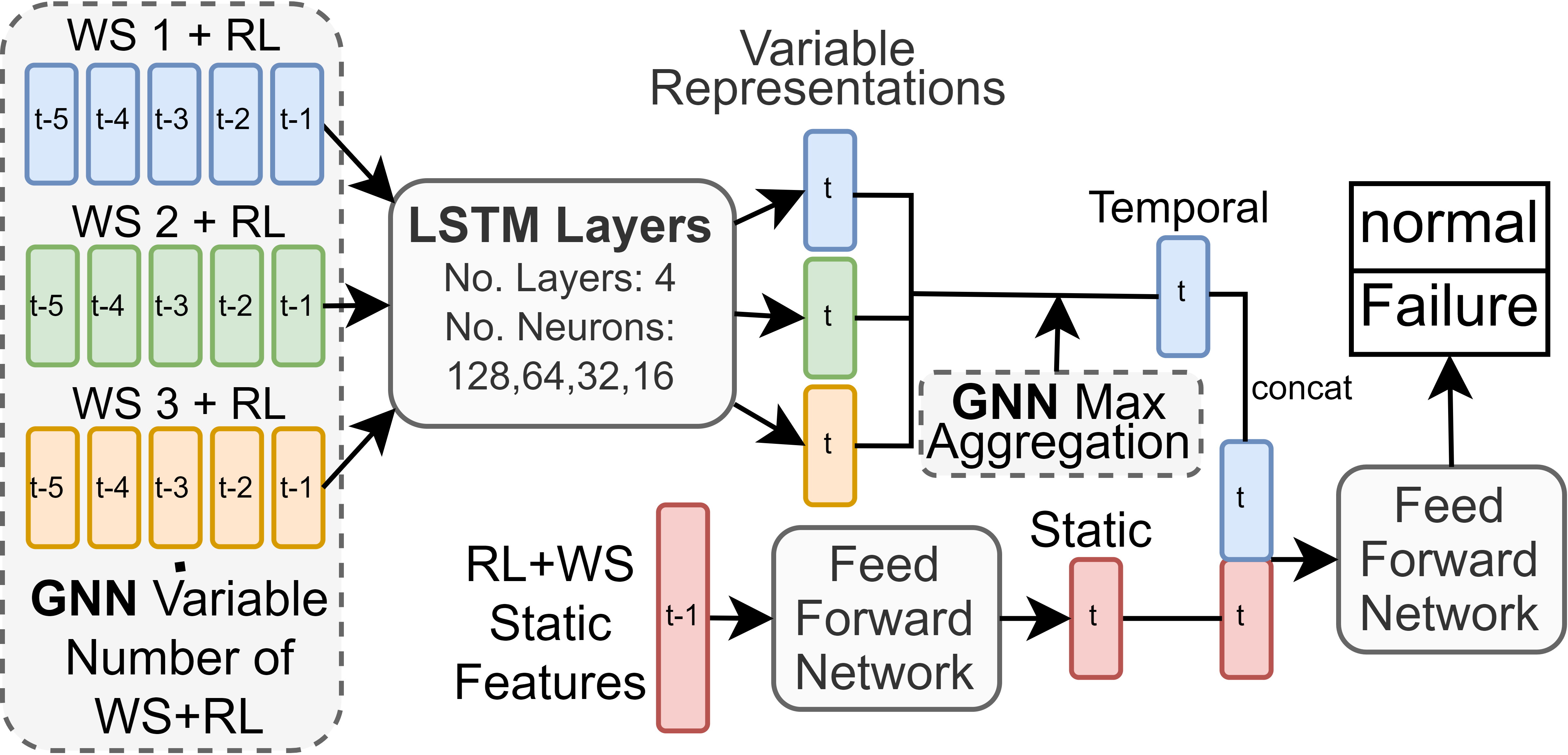}
    \caption{LSTM+ with the proposed GNN aggregation.}
    \label{fig:lstmp_g}
\end{figure}

\subsection{LSTM-AutoEncoder} 

Fig. \ref{fig:lstmae} and Fig. \ref{fig:gnnlstmae} present the LSTM-Autoencoder and its integration with the GNN aggregation, respectively. The LSTM-Autoencoder is only trained on normal links to encode the input sequence to a latent representation and decode it back to the output sequence \cite{said2020network}.  The encoder and decoder LSTM consists of two LSTM layers with decreasing (from 32 and 24) and increasing (24 and 32) number of neurons.

\begin{figure}[h]
    \centering
    \includegraphics[width=0.46\textwidth]{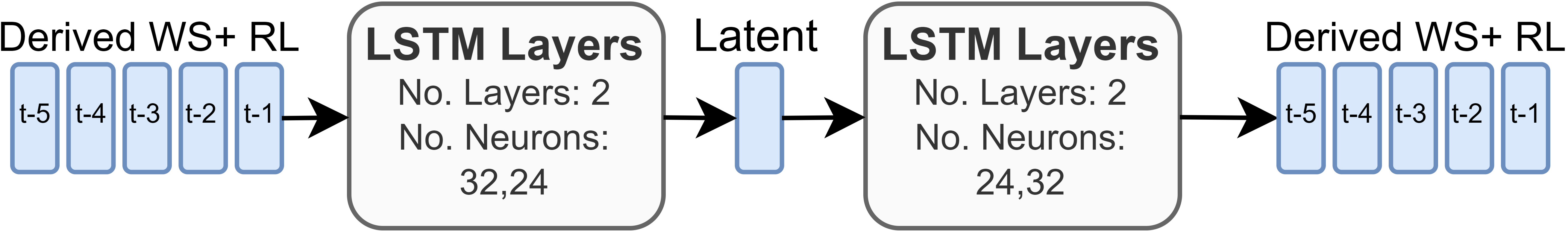}
    \caption{LSTM-Autoencoder architecture.}
    \label{fig:lstmae}
\end{figure}

We also augment the LSTM-Autoencoder architecture with our generalized graph aggregation method to show performance improvement. We introduce the variable number of weather station input handling and max aggregation from GenTrap to introduce LSTM-Autoencoder with GNN aggregation architecture as shown in Fig. \ref{fig:gnnlstmae}. This augmented network encodes input sequences to one latent vector and then repeats the vector $n$ times, where $n$ is the number of weather stations for the current batch. After this step, the repeated vectors are passed through a decoder network (like LSTM Decoder) to create output sequences similar to the input. 

\begin{figure}[h]
    \centering
    \includegraphics[width=0.46\textwidth]{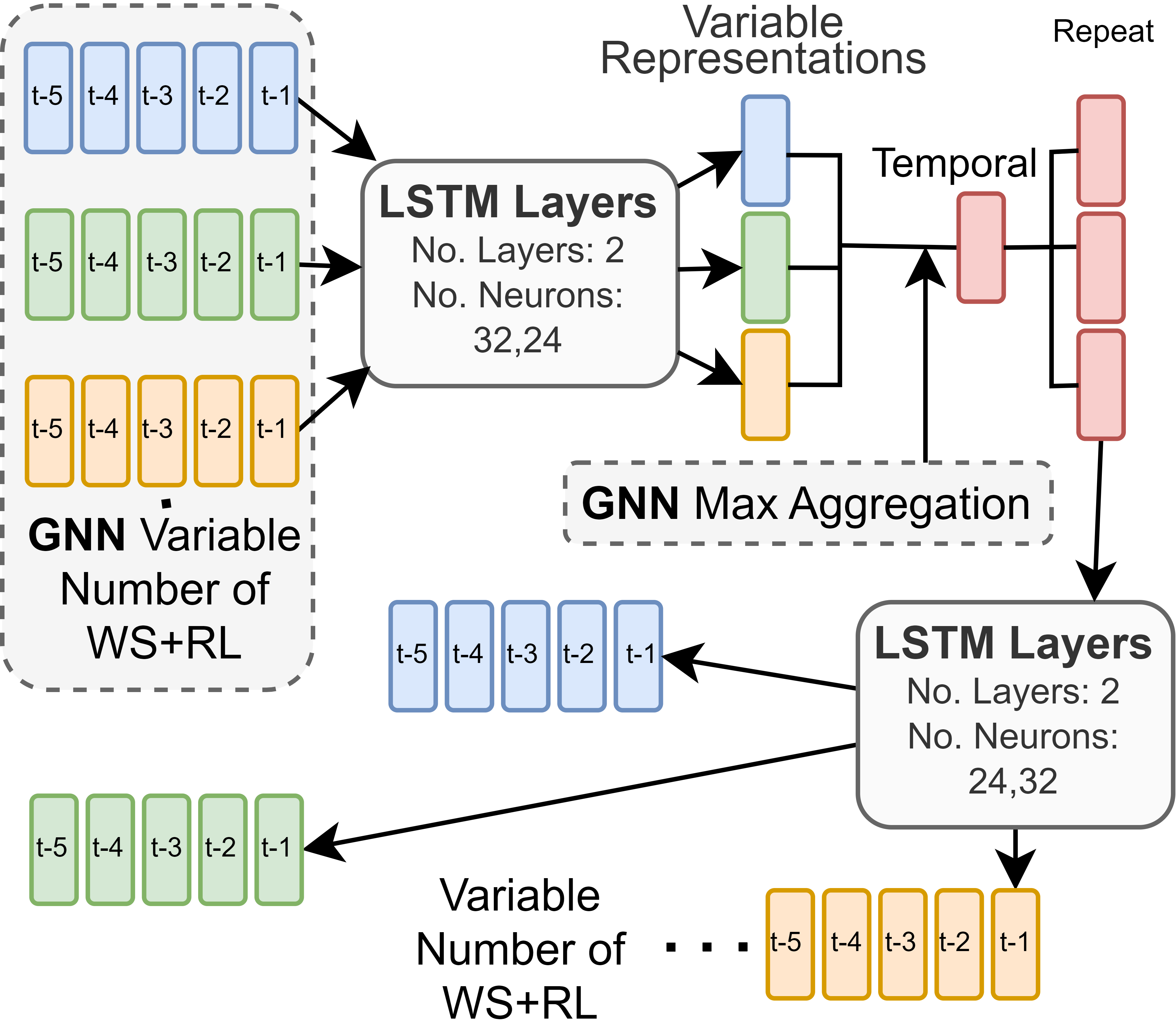}
    \caption{LSTM Autoencoder with proposed GNN aggregation.}
    \label{fig:gnnlstmae}
\end{figure}


\section{Radio Link Failure Prediction}


This section presents the link failure prediction workflow (Fig.~\ref{fig:workflow}) deploying the above learning models. The process consists of three components: data preprocessing, model training and validation, and model testing. In brief, the data preprocessing step consists of cleaning raw data, correlating links with weather stations, handling missing values, encoding categorical features, and performing a time series split. Then, the next part focuses on training and validating the existing LSTM+ and LSTM-autoencoder models along with the proposed GenTrap. Lastly, we test the performance of these approaches on unseen real-world link KPIs and weather observations, which is presented in Section~\ref{sec:evaluation}. 

\begin{figure}[h]
    \centering
    \includegraphics[width=0.46\textwidth]{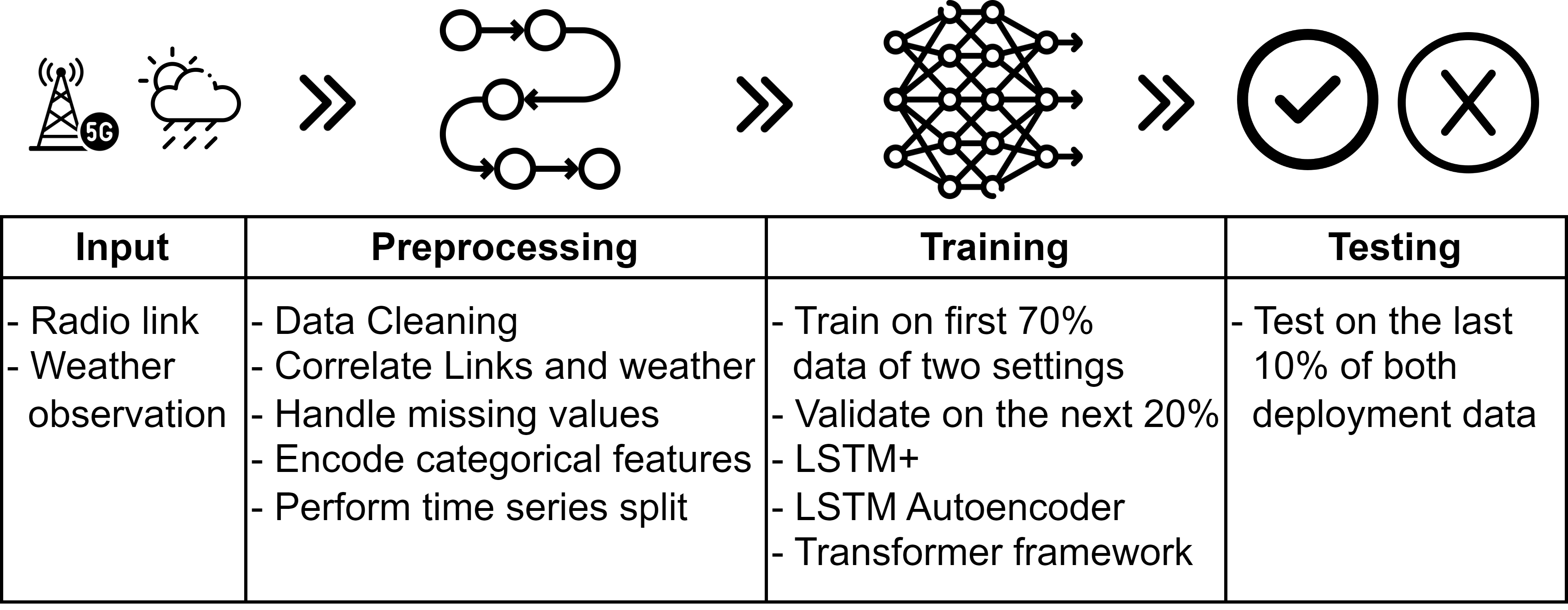}
    \caption{Link failure prediction workflow. }
    \label{fig:workflow}
\end{figure}

\subsection{Data Preprocessing}
The data preprocessing consists of the following steps.

\textbf{Data preparation.} The effectiveness, precision, and intricacy of machine learning tasks are significantly influenced by calibrating the training data \cite{gupta2021data}. Our initial investigation revealed that there are inconsistent values in weather station and radio link data (e.g., unexpected string values both in radio and weather data). These inconsistencies lead to erroneous or impossible data transformation for the subsequent steps, e.g., casting features to proper data types. Thus, we first tackle these inconsistencies, e.g., by removing the data samples if a numerical feature contained unexpected string values. After handling inconsistent values, we cast all numerical and categorical features to the floating and the string data type, respectively. In this problem, we consider daily data for both radio links and weather stations. 

\textbf{Real weather data alignment.} Our dataset has data from different entities (e.g., weather stations and their observations, radio sites and their link performance data). In order to merge weather observations with radio link KPIs, their temporal frequencies need to be maintained. Radio site KPIs and real weather realizations are collected in the chosen dataset over daily and hourly time intervals, respectively. We use the standard mean aggregation \cite{AKTAS2022107742} to transform hourly realizations into daily weather data to align historical weather realizations with radio link KPIs.

\textbf{Data imputation.} The majority of statistical and machine learning algorithms lack robustness in handling missing values, thereby being susceptible to the impact of incomplete data \cite{jadhav2019comparison}. We calculate the percentage of missing values for each feature in our dataset. Some features from historical radio link KPIs and real weather station data have a high percentage of missing values. We use a simple heuristic of dropping features with missing values of 20\% or higher. Also, some numerical features suffer from missing segments over time, but the data can be reliably interpolated if the percentage of missing values is under 15\% \cite{kreindler2016effects}. Thus, we deploy time series linear interpolation to impute missing numerical KPIs and historical weather observations \cite{pratama2016review}. 

\textbf{Data Merging.} We need to use historical KPIs and weather data to predict following-day link failure. Thus, we append a label column in the KPIs table, representing the next-day link status. Also, each radio site can have multiple links, so we merge the KPI features with the corresponding site features by matching the site id. Weather station features are also merged with weather observation data similarly.

\textbf{Tackling data imbalance.} 
We use the weighted cross-entropy loss function to tackle the data imbalance, which incorporates prior probabilities into a cost-sensitive cross-entropy error function. Unlike traditional cross-entropy, this weighted approach accounts for the imbalanced nature of the data, giving a larger influence to the majority class while minimizing overall error. The loss function puts the prior minority to majority class ratio $\lambda$ (0.003 for rural and 0.0006 for urban) into the regular cross entropy (Eq.~\ref{eq:weighted}). In rural deployment, this ensures that both classes have an equal influence because when $y=0$ for a non-failure instance, the remaining term $(1-y^i)\log(1-\hat{y}^i)$ only contributes $\lambda=0.3$ percent to the loss. Similarly, when $y=1$ for a failure instance, the remaining term $-y^i\log(\hat{y}^i)$ contributes $(1-\lambda)=99.7$ percent to the loss.

\textbf{Time series split.} 
Cross-validation (CV), a widely adopted method for assessing algorithm generalizability in classification and regression, has been extensively studied by researchers \cite{berrar2019cross}. Our dataset contains time series numerical values both for radio link KPI features and weather station observations \cite{AKTAS2022107742}. In the case of time series data, where the underlying process evolves over time, this can undermine the fundamental assumptions of cross-validation, which assumes that the data is independent and identically distributed. The temporal nature of time series data introduces dependencies and patterns that must be appropriately accounted for in the evaluation process \cite{bergmeir2012use}. Therefore, we use \textit{rolling origin} \cite{kreindler2016effects} method to compare our framework with previous works. 

The evaluation involves sequentially moving values from the validation set to the training set while changing the forecast origin accordingly. This way, folds with increasingly more train data are produced. It is also known as \textit{n-step-ahead} evaluation, where $n$ is the forecast horizon. The approach fits our use case because RLF prediction systems will be retrained or fine-tuned in the real world with new data as they become available. We create the 5 folds by first sorting the data across time and splitting them into the first 70\% train, the next 20\% validation and the last 10\% test set to create the fold with the largest training data. For instance, in the urban deployment - data ranging from January 2019 to December 2020 - this results in train, validation, and test sets containing January 2019 to April 2020, May 2020 to September 2020, and October 2020 to December 2020 data, respectively. This completes the first fold, and subsequent folds are created by offsetting the splits by 10\%. Thus, the second fold would contain the first 60\% as the train, the next 70\%-80\% as the validation, and the next 80\%-90\% as the test data. (Fig. \ref{fig:time_series}). Similarly, we create the rest of the folds for both deployments. 

\begin{figure}[h]
    \centering
    \includegraphics[width=3.4in]{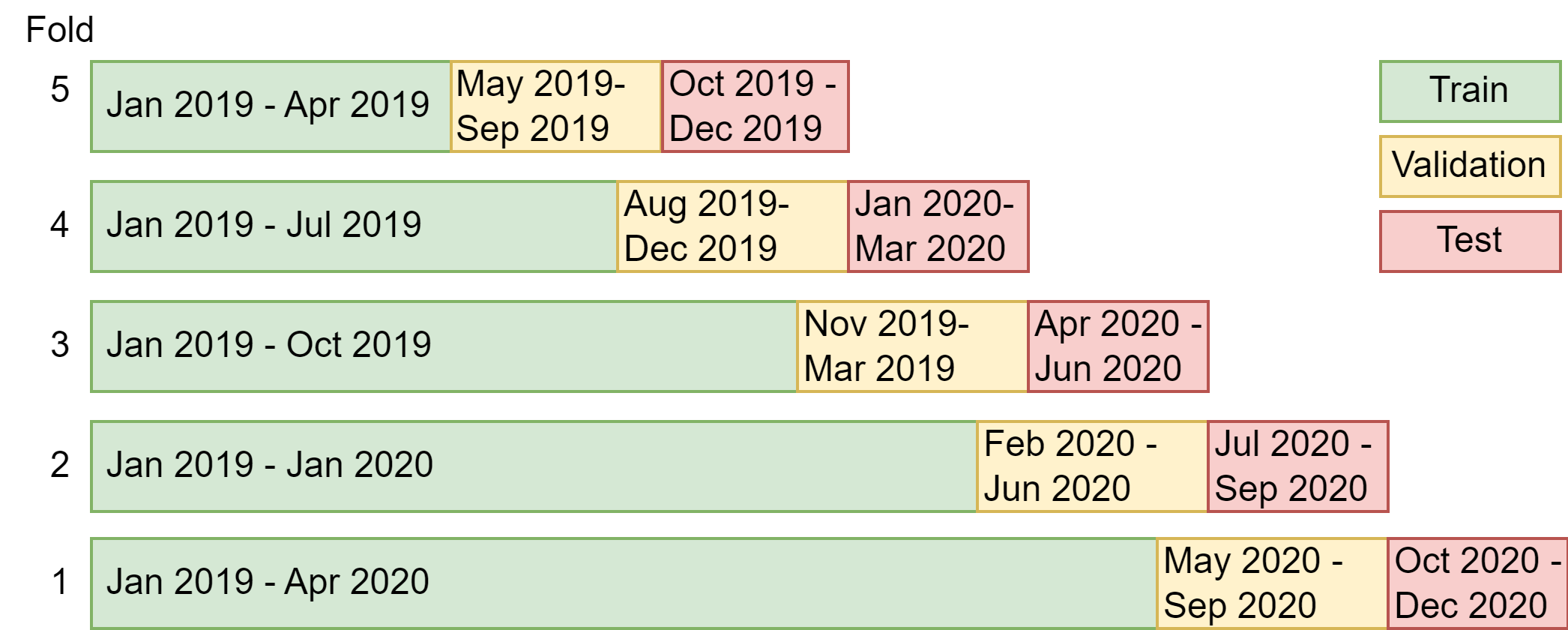}
    \caption{Rolling origin data splits for urban deployment.}
    \vspace{-0.2cm}
    \label{fig:time_series}
\end{figure}

\subsection{Model Training}

This section describes the model training of the proposed framework (GenTrap), existing state-of-the-art models (LSTM+ and LSTM-autoencoder), and existing models augmented with the learnable weather effect module. We first present the same model training parameters for all models and then present model-specific details for GenTrap, LSTM+, and LSTM-Autoencoder.

\textbf{Common Training Parameters.} We noticed that extremely low minority-to-majority class ratios (0.003 for rural and 0.0006 for urban) lead to unstable loss curves when a small batch size (e.g., 32 or 64) is used. Thus, we use a batch size of $1024$ for rural deployment and $6000$ for urban deployment to ensure that each batch has at least $2$ link failure events on average, which provided stability in model training. We predict the link failure probability for each day while using previous $5$ days' KPIs as that offers the best prediction performance \cite{Islam2022}). We use the same optimizer and loss function in all models.

\textbf{GenTrap Training.} We consider only the first $n$ $WS+RL(5,17)$ tensors in each batch of the model training, where $n=1$ to $3$. Thus, the same radio link can be associated with its $n$ closest weather stations in different epochs. This also works as a data augmentation scheme that increases the generalization capability of the model. During the inference, we remove the augmentation step and set $n=3$ to provide all surrounding weather station context to a link.

\textbf{LSTM+ Model Training.} We implement the LSTM+ approach introduced in \cite{AKTAS2022107742} to compare with our GenTrap framework. We follow the same data pre-processing steps as mentioned above with one difference for comparison purposes; we calculate derived features (mean, minimum, maximum, standard deviation of the $7$ weather station features) for each radio link from its $3$ closest weather station following \cite{AKTAS2022107742}. We use the $3$ closest weather stations for a fair comparison across the different architectures. We also augment LSTM+ architecture with our generalized graph aggregation method to measure the performance improvement gained by our framework. We only add the GNN aggregation module to the existing LSTM+ while keeping every other parameter the same.

\textbf{LSTM-Autoencoder Model Training.} Finally, we implement the LSTM-Autoencoder approach introduced in \cite{Islam2022} to compare it with GenTrap. Only the normal radio links are used to train the encoder-decoder LSTM network that captures feature dependencies of the normal scenario. During validation and testing (where we have both failure and normal link data), the trained model will struggle to recreate the input sequence for failed links because of the different input distributions of features during the model training. This way, when the reconstruction error is high, we can consider them predicted failure cases. Similarly, the GNN aggregation module is integrated into the LSTM-Autoencoder. The one key difference is the repeat step after performing max aggregation. This ensures the reconstructed sequences for each weather station and radio link pairs are decoded from the latent representation vector captured by the GNN aggregation module.

\section{Evaluation over Real-World data} \label{sec:evaluation}

This section presents the evaluation results of GenTrap over two real-world datasets: urban and rural. We also implement and evaluate the existing LSTM+ and LSTM-autoencoder models on the same datasets for a fair comparison. Next, we illustrate the benefits of integrating the learnable GNN aggregation module into existing models to boost their performance. Finally, we evaluate the generalization capability of GenTrap.    

\subsection{Performance Metrics and Evaluation Setup}

We evaluate the performance of different approaches using three metrics: precision, recall, and F1-score. For each approach, we first calculate true positives (TP), true negatives (TN), false positives (FP), and false negatives (FN) cases for both failure and non-failure events. \textit{True positives} are those failures in the test dataset that are correctly predicted as failures, while \textit{true negatives} are those non-failure events that are correctly predicted as non-failures. \textit{False positives} are the non-failures that are predicted as failures, while \textit{false negatives} are those failures that are predicted as non-failures. We then calculate the metrics for both failure and non-failure classes as follows: $Precision = \frac{TP}{TP+FP}, Recall = \frac{TP}{TP+FN},$ and $F1score = \frac{2×Precision×Recall}{Precision+Recall}$. We report the average of failure and non-failure precision, recall, and F1-score.

We perform all the experiments on a machine with Intel(R) Xeon(R) Silver 4210R CPU @ 2.40GHz, 32 GB memory, and Nvidia Quadro RTX 8000 GPU with 50GB VRAM. The OS and GPU versions were Ubuntu 20.04.6 LTS and CUDA 11.7, respectively.

\begin{table*}
\centering
\caption{The performance comparison of GenTrap.}
\label{tab:result}
\begin{tabular}{ | m{1.6cm} | m{0.6cm} m{0.6cm} m{0.6cm} | m{0.6cm} m{0.6cm} m{0.6cm} | m{0.6cm} m{0.6cm} m{0.6cm} | m{0.6cm} m{0.6cm} m{0.6cm} | m{0.6cm} m{0.6cm} m{0.6cm} | } 
  \hline
  \multicolumn{16}{|c|}{Rural} \\
  \hline
  \multicolumn{1}{|c|}{} & \multicolumn{3}{c|}{Nov-Dec 2019} & \multicolumn{3}{c|}{Oct-Nov 2019} & \multicolumn{3}{c|}{Sep-Oct 2019} & \multicolumn{3}{c|}{Aug-Sep 2019} & \multicolumn{3}{c|}{Jul-Aug 2019}\\
  \hline
  Approach & Pre-cision & Recall & F1-Score & Pre-cision & Recall & F1-Score & Pre-cision & Recall & F1-Score & Pre-cision & Recall & F1-Score & Pre-cision & Recall & F1-Score\\ 
  \hline
  GenTrap & \textbf{0.9994} & \textbf{0.8600} & \textbf{0.9183} & \textbf{0.9775} & 0.913 & \textbf{0.9431} & \textbf{0.9622} & 0.9758 & \textbf{0.9689} & \textbf{0.8571} & \textbf{0.9999} & \textbf{0.9166} & \textbf{0.9020} & \textbf{0.9115} & \textbf{0.9067} \\ 
  GenLSTM+ & 0.8456 & 0.8593 & 0.8523 & 0.8949 & \textbf{0.9275} & 0.9105 & 0.8571 & \textbf{0.9999} & 0.9166 & 0.8425 & 0.8466 & 0.8445 & 0.7600 & 0.7875 & 0.7731 \\ 
  LSTM+ & 0.7049 & 0.7581 & 0.7049 & 0.7713 & 0.7973 & 0.7837 & 0.8172 & 0.6687 & 0.7198 & 0.5881 & 0.7993 & 0.6359 & 0.6634 & 0.8451 & 0.7214 \\ 
  GNNLSTMAE & 0.6860 & 0.6192 & 0.6452 & 0.5092 & 0.5336 & 0.5138 & 0.6520 & 0.7128 & 0.6771 & 0.5052 & 0.6909 & 0.5103 & 0.4987 & 0.5000 & 0.4994 \\ 
  LSTMAE & 0.6650 & 0.5793 & 0.6070 & 0.5241 & 0.5455 & 0.5314 & 0.5790 & 0.5255 & 0.5377 & 0.5033 & 0.5980 & 0.5032 & 0.4998 & 0.4992 & 0.4994 \\ 
  \hline
\end{tabular}
\end{table*}

\begin{table*}
\centering
\begin{tabular}{ | m{1.3cm} | m{0.6cm} m{0.6cm} m{0.6cm} | m{0.6cm} m{0.6cm} m{0.6cm} | m{0.6cm} m{0.6cm} m{0.6cm} | m{0.6cm} m{0.6cm} m{0.6cm} | m{0.6cm} m{0.6cm} m{0.6cm} | } 
  \hline
  \multicolumn{16}{|c|}{Urban} \\
  \hline
  \multicolumn{1}{|c|}{} & \multicolumn{3}{c|}{Oct-Dec 2020} & \multicolumn{3}{c|}{Aug-Oct 2020} & \multicolumn{3}{c|}{Jun-Aug 2020} & \multicolumn{3}{c|}{Apr-Jun 2020} & \multicolumn{3}{c|}{Feb-Apr 2020}\\
  \hline
  Approach & Pre-cision & Recall & F1-Score & Pre-cision & Recall & F1-Score & Pre-cision & Recall & F1-Score & Pre-cision & Recall & F1-Score & Pre-cision & Recall & F1-Score\\ 
  \hline
  GenTrap & \textbf{0.8999} & \textbf{0.9799} & \textbf{0.9363} & 0.7383 & 0.8404 & \textbf{0.7803} & 0.6693 & \textbf{0.7378} & \textbf{0.6978} & \textbf{0.6734} & \textbf{0.8694} & \textbf{0.7360} & \textbf{0.8332} & \textbf{0.7856} & \textbf{0.8076} \\ 
  GenLSTM+ & 0.7544 & 0.9197 & 0.8168 & 0.6711 & \textbf{0.9061} & 0.7407 & \textbf{0.7270} & 0.6189 & 0.6560 & 0.6025 & 0.8476 & 0.6583 & 0.6599 & 0.6903 & 0.6738 \\ 
  LSTM+ & 0.7247 & 0.8347 & 0.7688 & \textbf{0.7527} & 0.7361 & 0.7441 & 0.5414 & 0.5057 & 0.5100 & 0.5822 & 0.7824 & 0.6273 & 0.5940 & 0.6902 & 0.6257 \\ 
  \hline
\end{tabular}
\end{table*}

\subsection{Performance comparison of different models}

Table \ref{tab:result} presents the performance of GenTrap along with LSTM+, LSTM-autoencoder, and their GNN aggregation-capable variants. We report F1 scores with their corresponding precisions and recalls on the 5-fold test data while predicting radio link failures for the following days of chosen days. The results confirm that GenTrap significantly and consistently outperforms all existing approaches with the best F1 scores of 0.93 and 0.79 for rural and urban deployment, respectively.  

The inferior performance of LSTM+ and LSTM Autoencoder can be attributed to two factors. First, these existing models deployed heuristics instead of learning the weather station association using GNN. Secondly, LSTM considers equal weights for all previous day data. It does not have any internal mechanism to give more priority to important days, e.g., feature values from recent days or important weather events. Another shortcoming of LSTM is its context window, which is limited to the previous context, thus facing the issue of vanishing gradients. This constraint can be a limitation when capturing complex dependencies with a long sequence span. 
On the other hand, GenTrap deploys GNN-based node aggregation for efficient spatial context capture. Furthermore, the transformer uses a self-attention mechanism to focus on the most relevant elements of the input sequence. Also, it has a larger context window to better understand the relationships between feature values that are far apart in the time sequence. These advantages from GNN aggregation and transformer time series encoding lead to GenTrap's superior performance.


Fig.~\ref{fig:label} presents the distribution and variability of F1-scores of different models using box and whiskers plots. The box represents the interquartile range (IQR), which contains the middle 50\% of the data. GenTrap scores are more concentrated in the middle with less variability, while scores from other approaches are more spread out. Having a lower variability makes GenTrap more reliable in real-world applications than previous approaches like LSTM+ and LSTM Autoencoder, which have greater variability. In the LSTM+ plot, we also observe outliers that lie beyond the whiskers, making this approach less credible.

We also perform a One-way ANOVA test to determine whether the GenTrap results are statistically significant. We achieve a $p$ value of 0.003, which proves the result is statistically significant at the 0.05 level.


\begin{figure}[h]
    \centering
    \includegraphics[width=0.4\textwidth]{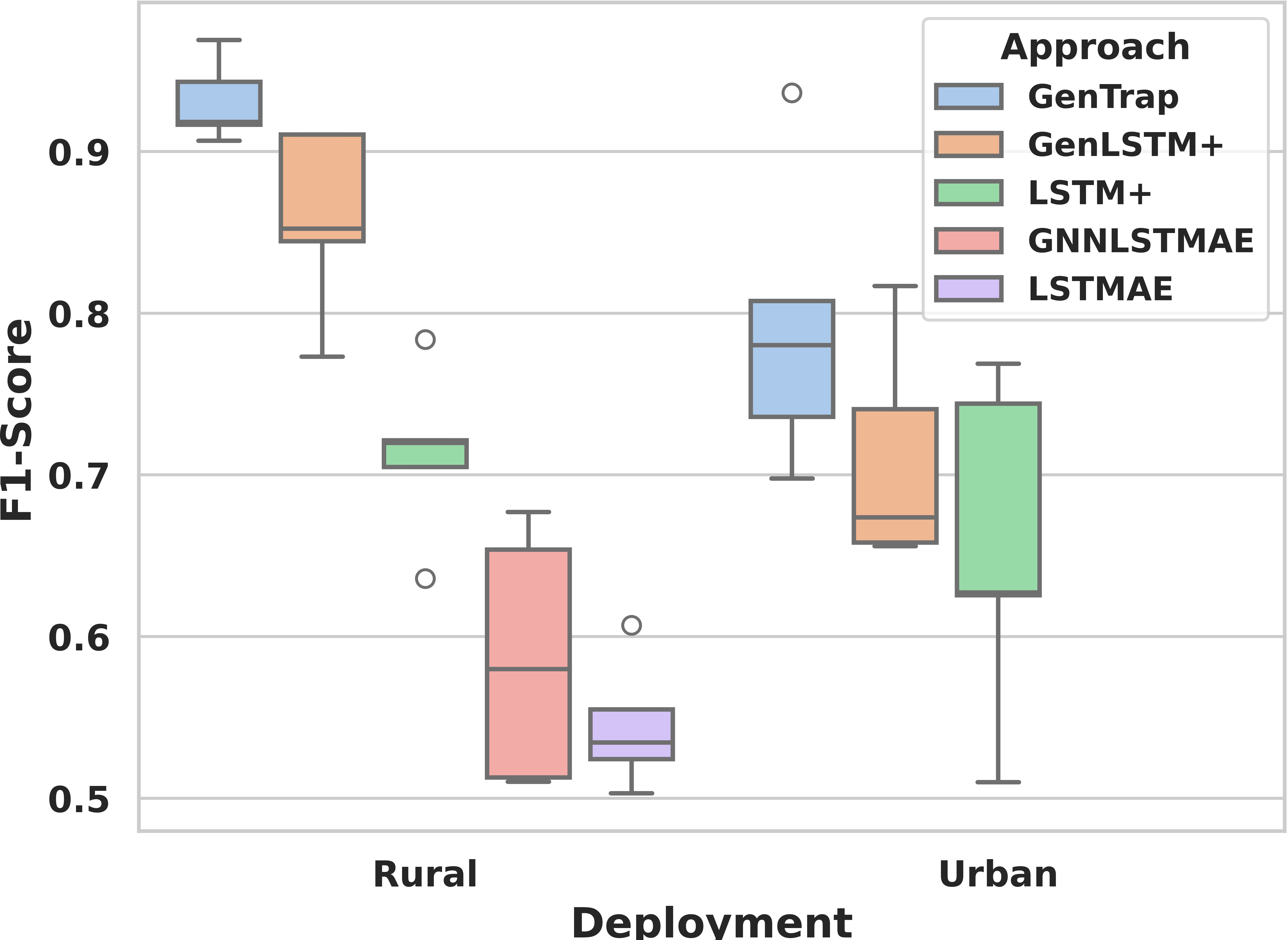}
    \caption{Distribution and variability of F1-scores of different models.}
    \label{fig:label}
\end{figure}


\subsection{Performance improvement using GNN aggregation}

The purpose of this evaluation is to show the benefits of having a learnable spatial context capture ability using GNN aggregation compared to existing heuristic-based context capture. Table~\ref{tab:result} presents the performance of GenLSTM+ and GenLSTMAE, i.e., LSTM+ and LSTM-Autoencoder augmented with GNN aggregation. GenLSTM+ significantly outperforms the heuristic-based LSTM+. In particular, GenLSTM+ offers an F1 score of 0.85 and 0.70 in rural and urban, respectively, compared to 0.71 and 0.65 in LSTM+. In the case of GenLSTMAE, we observe a similar improvement, i.e., an increase of F1 score from 0.54 to 0.58 on average - across the data splits for the rural deployment. Note that the LSTM-Autoencoder result of 0.60 has a worse performance than the reported one from previous work \cite{Islam2022}. We believe this is due to our consideration of the scalability score as a numerical feature instead of a categorical one. Also, the test dataset in the previous work included more failure events from approximately 6 months instead of 2 months of data. Furthermore, both GNNLSTMAE and LSTMAE perform like a random classifier (achieving an F1 score of only 0.4994) for the Jul-Aug 2019 split in rural deployment. This is likely due to the Autoencoder-based architecture's inability to extract information from a comparatively smaller data split during training and validation. Finally, we observe that LSTMAE models do not fit the large datasets from the urban deployment; thus, we report the performance from rural deployment and show the enhancement in the case of GNNLSTMAE.  


The LSTM+ and LSTM-Autoencoder used $k$ nearest weather stations and calculated an optimal distance from a radio link. These methods use heuristic-based weather station associations instead of dynamically learning the spatial context from surrounding weather stations. Thus, they suffer from worse performance as $k$ nearest and optimal distance methods must be tuned for any changes in the topology as these are susceptible to outliers. These methods also give the same weight to all associated weather stations, where the closer weather stations may have more influence. On the other hand, we introduce a GNN aggregation step which can benefit existing architectures, such as LSTM+ and LSTM-Autoencoder, to dynamically learn which weather stations to focus on for each link and assign weights accordingly; hence, obtain better performance and generalization ability. 


\subsection{Generalization performance of GenTrap}

This last evaluation focuses on GenTrap's generalization ability to show its application in modern 5G RAN. Usually, providers selectively turn off radio stations for resource (e.g., energy) savings as the traffic demand across base stations can vary according to their locations (urban vs. rural) and time of the day (working vs. after working hours) \cite{ma2022optimal}\cite{umar2020base}. Thus, a prediction model must be generalized, i.e., capable of adapting with dynamically changing links. Another benefit of such generalization is saving computational resources to train the model with a subset of links instead of the entire set.  

In this evaluation, we train GenTrap and LSTM+ models on different fractions of the links of the given topology while testing the entire topology to understand how well the models generalize on unseen links. This gives us a measure of how good the chosen models are in learning from a topology with fewer links and generalizing over a bigger topology with new links. Table \ref{tab:gen} presents the comparison of GenTrap with LSTM+ as it performs better than the LSTM-Autoencoder scheme. We take fractions 0.1 to 0.5 of the links from the rural deployment. GenTrap consistently outperforms LSTM+ in all fractions, with an average F1 score of 0.70 compared to 0.63. We also notice the improvement tends to be greater for larger fractions (0.3, 0.4, 0.5) with an average improvement from 0.65 to 0.76 than for smaller fractions (0.1, 0.2) with an average improvement from 0.59 to 0.61. This suggests that as the topology grows, we will benefit more from GenTrap. In only one data split (Aug-Oct 2020), LSTM+ performs better than GenTrap in Precision. This could be due to LSTM+ having a higher trade-off between precision and recall for that specific split.  

LSTM+ has no specific architectural component focused on generalization to unseen links. It calculates the derived features from $k$ nearest weather stations for each link, i.e., considers each link once. Our GNN-based variable weather station aggregation method allows GenTrap to introduce data augmentation by considering different numbers of weather stations for the same link during training time. We attribute the improvement in generalization due to the inherent data augmentation technique in GenTrap. 


\begin{table}
\centering
\caption{ Generalization comparison of GenTrap and LSTM+ for rural deployment. }
\label{tab:gen}
\begin{tabular}{|m{1cm}| m{0.6cm} m{0.6cm} m{0.7cm}|m{0.6cm} m{0.6cm} m{0.7cm}|} 
 \hline
 \multirow{2}{*}{\shortstack{Training\\link\\fraction}} & \multicolumn{3}{c|}{GenTrap} & \multicolumn{3}{c|}{LSTM+} \\ \cline{2-7}
 & Pre-cision & Recall & F1-score & Pre-cision & Recall & F1-score \\ 
 \hline
 0.5 & 0.7157 & 0.7587 & 0.7352 & 0.6840 & 0.6987 & 0.6910\\ 
 \hline
 0.4 & 0.8857 & 0.8396 & 0.8612 & 0.6024 & 0.7365 & 0.6423 \\ 
 \hline
 0.3 & 0.6480 & 0.7775 & 0.6927 & 0.6006 & 0.7561 & 0.6436 \\ 
 \hline
 0.2 & 0.6068 & 0.6578 & 0.6272 & 0.5655 & 0.6951 & 0.5969 \\ 
 \hline
 0.1 & 0.5867 & 0.6178 & 0.5998 & 0.5846 & 0.5981 & 0.5908 \\ 
 \hline
\end{tabular}
\end{table}

\subsection{Discussion}

This section discusses how GenTrap can be extended in future. 

\textbf{Enhancing GenTrap model.} This work demonstrated how a GNN aggregation and Transformer-based spatiotemporal context capture could lead to increased performance and generalization of a radio link failure prediction scheme. However, the proposed model can further be improved by incorporating recent advancements in pre-training transformers and transformer GNN. Following an unsupervised pre-training scheme similar to \cite{zerveas2021transformer,shi2021self}, GenTrap can benefit from performance enhancement over the current fully supervised learning. Similarly, the GNN aggregation module in GenTrap can be extended with a transformer encoder like \cite{dwivedi2020generalization,min2022transformer} to directly learn the aggregation function instead of using the max aggregation. We can also utilize the GNN aggregation to capture inter-base station effects such as interference. The same architecture principle can be applied to purely unsupervised approaches where the input consists of a variable number of weather stations for each radio link. We plan to explore these options in future.


\textbf{Improving data quality.} Our datasets have a meagre minority-to-majority class ratio. Because of that, a small volume of data (minority class) can penalize the model performance. The failure events present in the dataset may not capture all possible and essential cases due to the challenges of real-world data collection. We believe a reliable and truthful generation of synthetic failure events, with the help of simulations \cite{gkonis2020comprehensive} or generative models \cite{wen2020time}, can improve the data quality and thus increase the faithfulness of deep learning models.


\section{Conclusion}

5G RAN radio links can fail due to changes in weather conditions. A proactive RLF prediction system can improve user experience and save network operators' time, cost, and resources. Thus, we investigated the shortcomings of existing link failure prediction models and proposed a novel GNN aggregation and time-series transformer-based framework called GenTrap. It deploys GNN aggregation over a variable number of surrounding weather stations to capture spatial context while incorporating the transformer for temporal context. The evaluation of GenTrap over two real-world datasets confirmed its superiority over existing LSTM-based models. We also demonstrated that integrating the GNN aggregation into existing models could improve performance. Finally, we presented the generalization capability of GenTrap in the presence of unseen links. Thus, service providers can deploy GenTrap for predictive maintenance to support emerging IoT applications.    


\bibliographystyle{IEEEtran}
\bibliography{references}
\end{document}